\documentclass[conference]{IEEEtran}
\IEEEoverridecommandlockouts
\usepackage{cite}
\usepackage{amsmath,amssymb,amsfonts}
\usepackage{algorithmic}
\usepackage{graphicx}
\usepackage{textcomp}
\usepackage{xcolor}

\usepackage{dirtree}
\usepackage{ulem}
\usepackage{subfig}
\usepackage{graphicx}
\usepackage{verbatim}
\usepackage{subfig}
\usepackage{ulem}
\usepackage{dblfloatfix}
\usepackage{booktabs}
\usepackage{amsmath}
\usepackage{fancyhdr}
\usepackage{bbm}
\usepackage{enumitem}
\usepackage{minted}
\usepackage{balance}
\usepackage{fancyhdr}
\usepackage{multicol}
\usepackage{bbding}
\usepackage{svg}
\usepackage{multirow}
\usepackage{tablefootnote}
\usepackage{svg}
\usepackage{footmisc}
\usepackage{hyperref}
\usepackage{subfig}
\usepackage{listings}
\usepackage{hyperref}
\usepackage{eso-pic}

\def\BibTeX{{\rm B\kern-.05em{\sc i\kern-.025em b}\kern-.08em
    T\kern-.1667em\lower.7ex\hbox{E}\kern-.125emX}}

\newcommand{\cc}{{\sc CleverCatch}}
\newcommand{\pt}

\begin{document}

\AddToShipoutPictureBG*{%
  \AtPageUpperLeft{%
    \raisebox{-0.8cm}{%
      \makebox[\paperwidth]{%
        \centering
        \scriptsize Published in the Proceedings of the 2025 IEEE International Conference on Big Data (BigData 2025).
        DOI: 10.1109/BIGDATA66926.2025.11401453
      }%
    }%
  }%
}

\bstctlcite{IEEEexample:BSTcontrol}
\title{CleverCatch: A Knowledge-Guided \\ Weak Supervision Model for Fraud Detection 
}

\makeatletter
\renewcommand{\footnotesize}{\scriptsize} 
\setlength{\footskip}{20pt}               
\makeatother

\fancypagestyle{firstpage}{
    \fancyhf{}
    \renewcommand{\headrulewidth}{0pt}
    \renewcommand{\footrulewidth}{0pt}
    \fancyfoot[L]{%
        \raisebox{2pt}{\rule{\textwidth}{0.4pt}}\\[-4pt]%
        \footnotesize * Corresponding author%
    }
}

\author{ \IEEEauthorblockN{{\large \textbf{Amirhossein Mozafari$^*$}}}
\IEEEauthorblockA{{\large\textit{School of Computing Science}} \\
\textit{{\large Simon Fraser University}}\\
Burnaby, Canada \\
amozafar@sfu.ca}
\and
\IEEEauthorblockN{{\large\textbf{Kourosh Hashemi}}}
\IEEEauthorblockA{{\large \textit{School of Computing Science}} \\
\textit{{\large Simon Fraser University}}\\
Burnaby, Canada \\
kourosh\_hashemi@sfu.ca}
\and
\IEEEauthorblockN{\textbf{{\large Erfan Shafagh}}}
\IEEEauthorblockA{\textit{{\large School of Computing Science}} \\
\textit{{\large Simon Fraser University}}\\
Burnaby, Canada \\
erfan\_shafagh@sfu.ca}
\and
\hspace{0.7cm}
\IEEEauthorblockN{\textbf{{\large Soroush Motamedi}}}

\IEEEauthorblockA{
\hspace{0.7cm}\textit{ {\large School of Computing Science}} \\
\textit{
\hspace{0.7cm}{\large Simon Fraser University}}\\
\hspace{0.7cm}
{Burnaby, Canada} \\
\hspace{0.7cm}
soroush\_motamedi\_sedeh@sfu.ca}
\and
\IEEEauthorblockN{\textbf{{\large Azar Taheri Tayebi}}}
\IEEEauthorblockA{\textit{{\large Department of Computer Science}} \\
\textit{{\large Brock University}}\\
Ontario, Canada \\
at22gg@brocku.ca}
\and
\IEEEauthorblockN{\textbf{ {\large Mohammad A. Tayebi}}}
\IEEEauthorblockA{\textit{{\large School of Computing Science}} \\
\textit{{\large Simon Fraser University}}\\
Burnaby, Canada \\
tayebi@sfu.ca}
}
\vspace{-0.3cm}
\maketitle
\thispagestyle{firstpage}
\begin{abstract}

Healthcare fraud detection remains a critical challenge due to limited availability of labeled data, constantly evolving fraud tactics, and the high dimensionality of medical records. Traditional supervised methods are challenged by extreme label scarcity, while purely unsupervised approaches often fail to capture clinically meaningful anomalies. In this work, we introduce \cc, a knowledge-guided weak supervision model designed to detect fraudulent prescription behaviors with improved accuracy and interpretability. Our approach integrates structured domain expertise into a neural architecture that aligns rules and data samples within a shared embedding space. By training encoders jointly on synthetic data representing both compliance and violation, \cc~ learns soft rule embeddings that generalize to complex, real-world datasets. This hybrid design enables data-driven learning to be enhanced by domain-informed constraints, bridging the gap between expert heuristics and machine learning. Experiments on the large-scale real-world dataset demonstrate that \cc~ outperforms four state-of-the-art anomaly detection baselines, yielding average improvements of 1.3\% in AUC and 3.4\% in recall. Our ablation study further highlights the complementary role of expert rules, confirming the adaptability of the framework. The results suggest that embedding expert rules into the learning process not only improves detection accuracy but also increases transparency, offering an interpretable approach for high-stakes domains such as healthcare fraud detection.
\end{abstract}




\maketitle

Healthcare fraud remains a significant and costly issue in public insurance programs, with global losses estimated in the tens of billions annually \cite{national2018crossing, shrank2019waste, fbi_healthcare_fraud}. Around 7\% of worldwide health spending, approximately \$560 billion, is lost to fraud and corruption each year \cite{transparency2021health}. In the U.S., the National Health Care Anti-Fraud Association estimates that about 3\% of healthcare spending, or roughly \$300 billion, is lost to fraud annually \cite{nhcaa2021fraud}. Similarly, the Canadian Life and Health Insurance Association (CLHIA) estimates that 2\% to 10\% of healthcare dollars in North America are affected by fraud, indicating substantial losses in Canada as well \cite{clhia2023fraud}. These losses not only divert vast sums into the wrong hands but also reduce access to essential medical services for those in need. Thus, implementing an effective fraud detection system is crucial to protect the public's well-being.

One of the most complicated and difficult-to-spot forms of healthcare fraud involves prescription drug claims. In these cases, some providers take advantage of the reimbursement system by prescribing excessive, unnecessary, or chosen medications to maximize profit rather than patient need. What makes this even more challenging is that fraud isn’t static; schemes constantly evolve, and those behind them are quick to adjust their methods to stay one step ahead of oversight and regulation.

Existing research in healthcare fraud detection has primarily explored various machine learning methods to identify fraudulent patterns in claims data. Supervised learning techniques have been widely applied when labeled datasets are available \cite{bauder2016outlier, johnson2019medicare, pang2021deep, herland2018bigdata}. While these methods often demonstrate strong performance, they rely heavily on high-quality, accurately labeled data. This requirement is difficult to meet in real-world fraud scenarios, where confirmed cases are rare and labeling is both expensive and time-consuming. In response, unsupervised learning methods, particularly anomaly detection techniques, have been employed to identify atypical patterns that may indicate fraud without the need for labeled instances \cite{sadiq2017primm, suesserman2023procedure, kennedy2025unsupervised}. These approaches are especially valuable in the healthcare domain, where fraudulent behavior often appears as slight deviations from normative patterns.

More recently, weakly supervised learning has gained traction by leveraging partially labeled or noisy data to maintain a balance between the need for supervision and the lack of labeled data \cite{ruff2020deepsemisupervisedanomalydetection, pmlr-v80-ruff18a, pang2019deep, pang2023deep}. Knowledge-guided machine learning is another promising direction for anomaly detection tasks \cite{rao2021knowgnn, zhou2024knowgraph, visbeek2023dsc, li2024sefraud, pan2025huge, park2022das}. In this approach, domain-specific knowledge, such as expert-defined rules, is integrated into data-driven models to enhance both performance and interpretability. Despite the growing body of work in this area, there is relatively little research specifically focused on fraud detection. To address this gap, we propose \cc, a solution designed specifically to detect fraudulent behavior in prescription drug claim data.

To build our knowledge-guided fraud detection model, and given the limitations of our training data, we focus on two classes of knowledge-based rules derived from expert insights into prescription behavior. The first targets cost-preference anomalies, identifying physicians who consistently favor higher-cost drugs over clinically equivalent \cite{cheng2019network}, lower-cost alternatives. The second centers on opioid prescribing patterns, flagging unusually high reliance on opioids informed by prior research on opioid overuse and overprescribing \cite{zafari2019topic, guy2017vitalsigns}.

\cc, introduces a novel framework that integrates structured domain knowledge into a base anomaly detection model by embedding expert rules into a shared low-dimensional latent space. Unlike rigid rule-based or purely data-driven methods, it learns soft rule representations and their relationship to data, allowing flexible reasoning about rule satisfaction. The {\it Rule Encoder (RE)} and {\it Sample Encoder (SE)} are jointly trained on synthetic data representing both compliance and violations, enabling robust generalization to high-dimensional real-world data with limited labeled fraud examples. This embedding-based approach helps \cc~ capture subtle fraud patterns through geometric relationships between data and rule embeddings, enhancing interpretability and adaptability for complex domains like healthcare fraud detection. Moreover, the soft alignment (via optimal transport) of the embedded rules and data samples, makes \cc robust to a small number of incorrect rules (noisy rules), ensuring performance degrades smoothly and no single rule determines the outcome.\par
Our experiments on a large Medicare Part D dataset show that \cc\ consistently outperforms four state-of-the-art baseline models, achieving average improvements of 1.3\% in AUC and 3.4\% in recall across all comparisons. Notably, cost-preference rules drive the largest gains, while opioid-related rules offer complementary signals, enhancing the model’s ability to detect a broader range of fraud patterns often missed by purely data-driven approaches. These findings highlight the effectiveness of incorporating domain knowledge into fraud detection systems.

\noindent In summary, our key contributions are as follows:

\begin{itemize}[leftmargin=*, label=$\diamond$]
\item We propose a novel approach to healthcare fraud detection that integrates structured domain knowledge into a base anomaly detection model, advancing the state of the art. To the best of our knowledge, this is the first method of its kind. 

\item We define two sets of domain-informed rules focused on cost-preference and opioid prescribing behaviors. These rules reflect expert knowledge and address complex, evolving fraud scenarios in a principled and interpretable way.

\item Through experiments on large-scale, real-world healthcare data, we demonstrate that our model not only outperforms strong baseline methods, but also that the inclusion of expert-defined rules significantly enhances detection performance.
\end{itemize}

This paper is organized as follows: Section~\ref{sec:data} describes the dataset; Section~\ref{sec:domain} outlines domain-informed fraud scenarios; Section~\ref{sec:prelim} introduces preliminaries; Section~\ref{sec:methodology} details our method; Section~\ref{sec:experiments} presents results; Section~\ref{sec:related} reviews related work; and Section~\ref{sec:conclusion} concludes.

\begin{table}[t]
\centering
\caption{Summary Statistics of Medicare Part D Dataset  (2013–2023)}
\label{tab:basic_stats}
\begin{tabular}{@{}ll@{}}
\toprule
\textbf{Statistic} & \textbf{Value} \\
\midrule
Number of Unique Physicians (NPIs) & 1,635,865 \\
Number of Unique Generic Drugs & 2101 \\
Number of Unique Specialties & 280 \\
Average Unique Drugs Prescribed per Physician & 38 \\
Average Total Claims per Physician & 8654  \\ 
Average Total Cost per Physician & 905,539 USD \\
Number of Physicians Linked to Fraud (LEIE) & 2321 \\
Percentage of Fraud-Labeled Physicians & 0.14 percent \\
\bottomrule
\end{tabular}
\end{table}

\section{Data and Context Overview}\label{sec:data}
\label{sec:data}
\subsection{Data Characteristics}
\label{sec:dataCharacteristics}
We focus our study on the Medicare Part D dataset, which contains information on prescription drugs entered by physicians into an electronic medical record system for a given year. The dataset spans 10 years, from 2013 to 2023 and is publicly available on the Medicare \& Medicaid Services (CMS) website \cite{cmsPartD2025}. Each row in the dataset represents a unique combination of physician, specialty type, and drug name. The physician is identified by a unique National Provider Identifier (NPI), and a single physician may be associated with multiple specialty types. 


In addition to a drug’s brand and generic names, the dataset includes several drug-related metrics: total cost, total claim count, total number of beneficiaries, 30-day fill count, and total daily supply, each specific to the physician in a given year. 

To assign binary fraud labels, we use data from the List of Excluded Individuals and Entities (LEIE), maintained by the Office of Inspector General (OIG) and updated monthly \cite{leie2025}. This list, updated monthly, identifies physicians currently excluded from federal healthcare programs. By linking the NPIs from the LEIE to those in the Part D dataset, we identify fraud-labeled physicians. Note that the dataset is highly imbalanced, with fraud-labeled physicians representing only 0.14\% of the entire population. Table~\ref{tab:basic_stats} summarizes key statistics of the used datasets for our experimental evaluation.

\begin{figure}[t!]%
\centering
\subfloat[]	{
  \includegraphics[width=40mm, height=35mm]{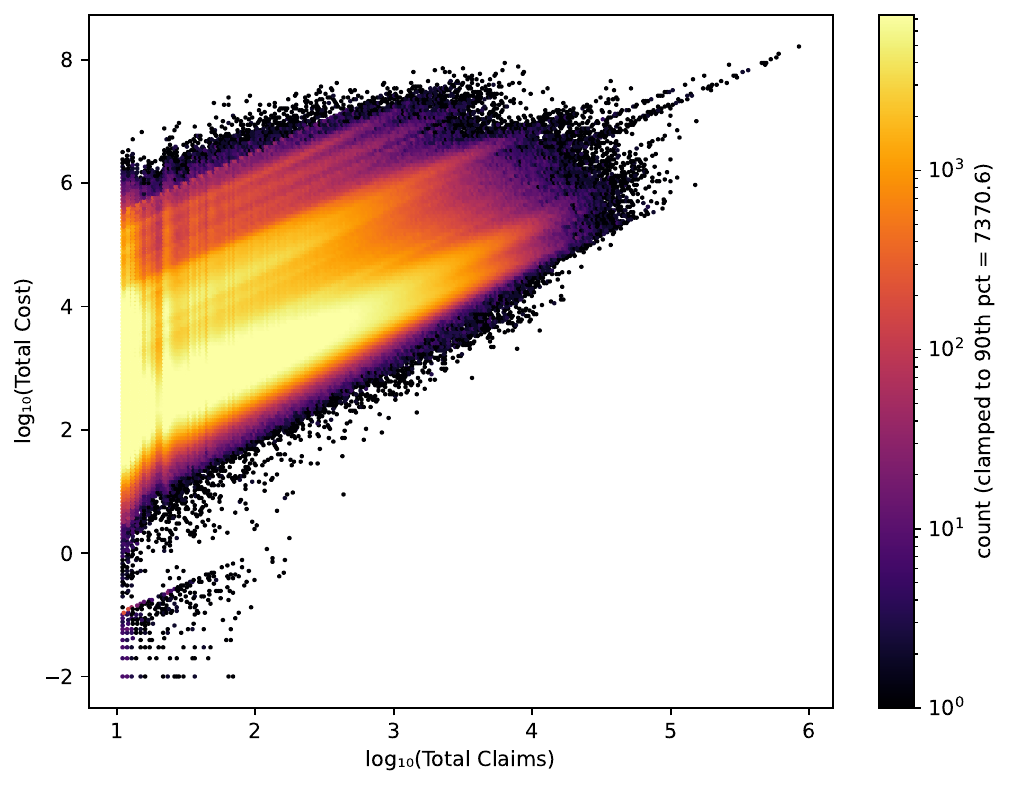}
    \label{fig:bnec}
}
\subfloat[]	{
  \centering
  \includegraphics[width=40mm, height=35mm]{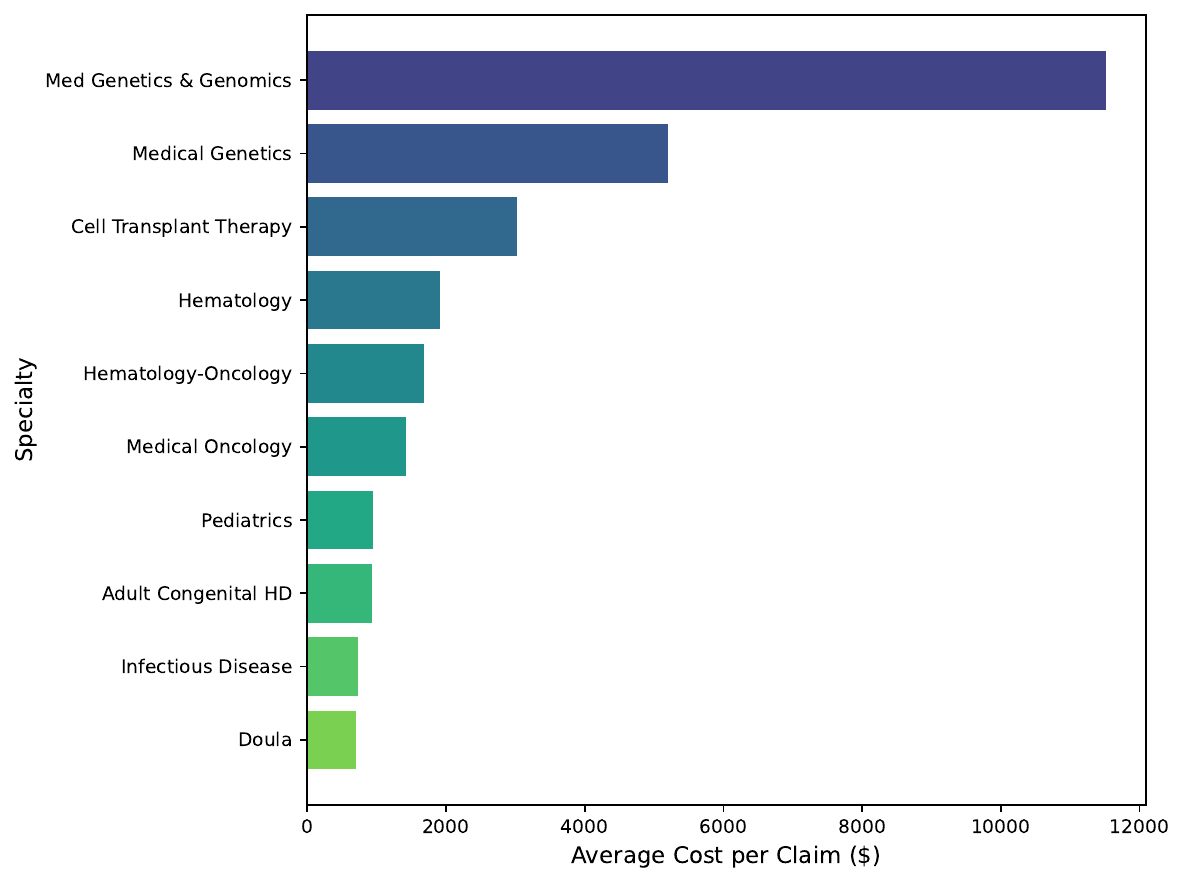}
  \label{fig:bner}
	}
	
		\subfloat[]	{
  \centering
  \includegraphics[width=40mm, height=35mm]{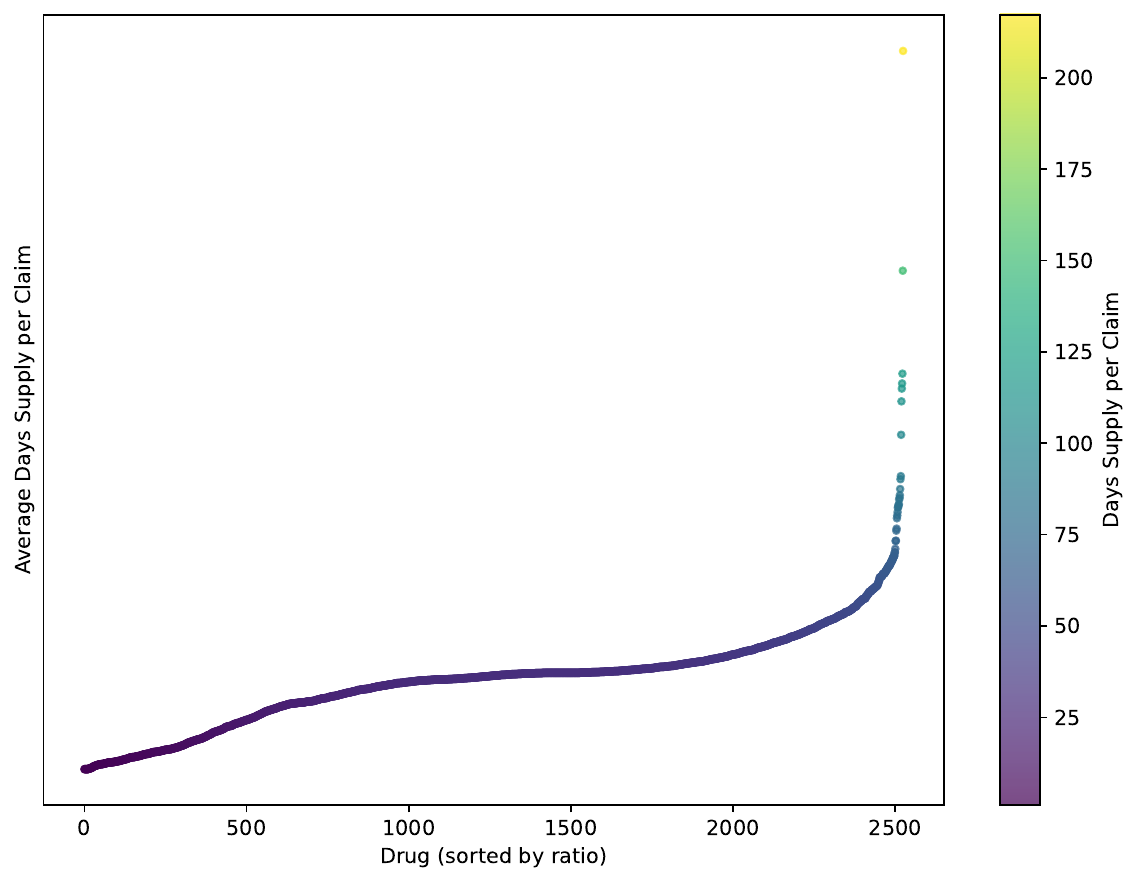}
  \label{fig:tfauto}
	}
	\subfloat[]	{
  \centering
  \includegraphics[width=40mm, height=35mm]{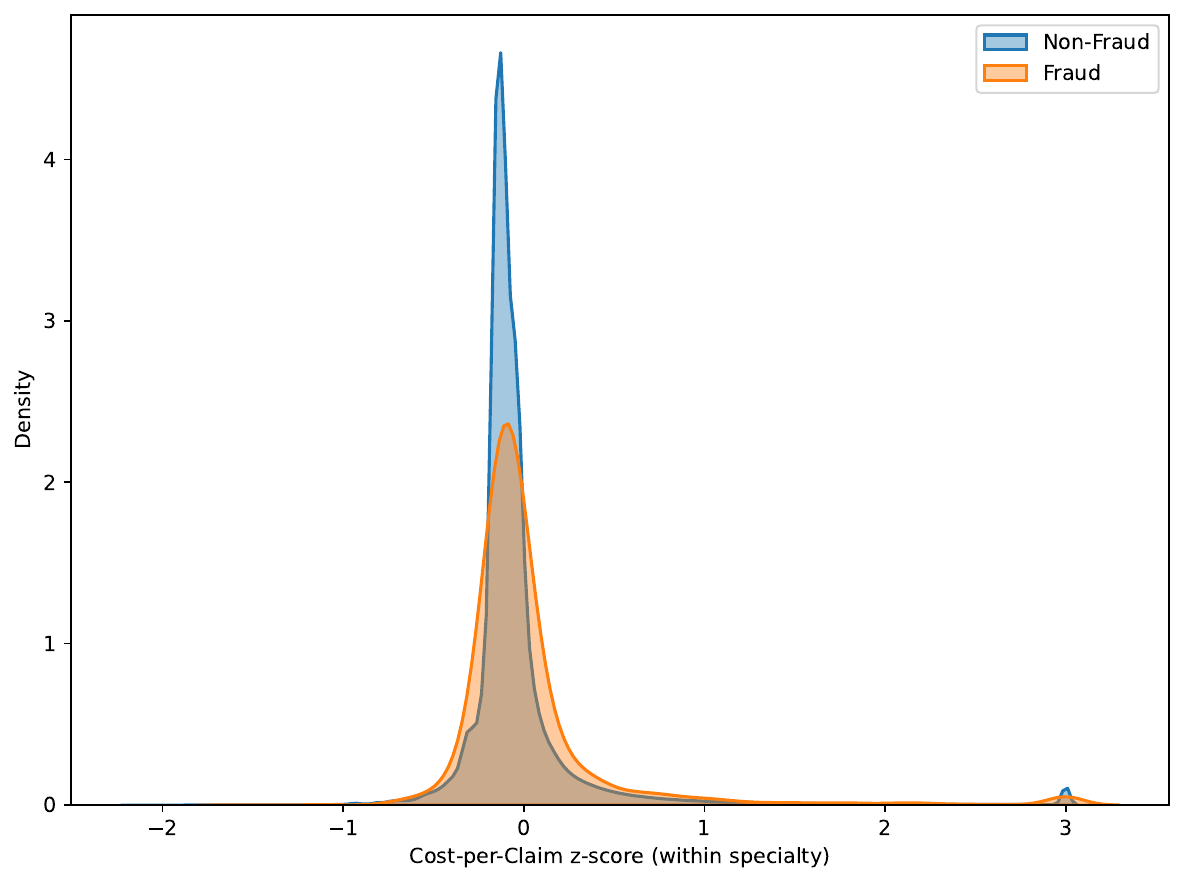}
  \label{fig:stauto}
	}
    \caption{Prescription behavior and fraud patterns: (a) Total cost vs. total claims for all physician-drug combinations; (b) Mean drug cost per claim by specialty; (c) Average daily supply per claim; (d)  Average cost per claim between fraud-labeled and non-fraud physicians.}
		\label{fig:VancouveCrimes}
\end{figure}

Figure \ref{fig:bnec} shows the total cost versus the total claim count for all drug-physician combinations. 
In log–log axes with 90th-percentile color clamping, most physician–drug pairs fall in a moderate cost–volume band, while a few lie as outliers at very low or very high extremes. Figure \ref{fig:bner} presents the total cost and total number of claims by physician specialty to examine differences across domains. As expected, specialties Medical Genetics and Genomics have the highest average costs. Figure \ref{fig:tfauto} displays the ratio of daily supply to total claims to assess prescribing intensity per drug. The right-skewed distribution indicates that most drugs have low to moderate intensity, with a few exhibiting significantly higher values. In Figure \ref{fig:stauto} fraud‐flagged prescribers (orange) show a noticeably wider, flatter z-score distribution compared to non-fraud physicians (blue), indicating they vary more around their specialty’s average cost per claim. Moreover, the orange curve is shifted slightly to the right of zero, revealing that fraud-flagged physicians tend to have higher cost-per-claim relative to their peers.

\subsection{Fraud Scenario}\label{sec:domain}
Fraudulent behavior in medical prescription data often reflects not just statistical outliers but also discernible patterns rooted in domain-specific knowledge. Such patterns can frequently be expressed as rules, hypotheses formulated by experts regarding how fraud might manifest in practice. These rules may capture unethical prescription tendencies, unusual drug choices, or prescribing practices inconsistent with clinical norms \cite{sun2020medical, zare2016improving}.The validity of any fraud scenario is inherently constrained by the available data, and assumptions about physician or patient behavior must be evaluated against the attributes the dataset supports. When key elements such as patient histories or longitudinal context are missing, certain fraud hypotheses cannot be reliably tested. Accordingly, our scenario design acknowledges these limitations and defines rules based only on what can be reasonably observed within the training data. Many fraud scenarios in healthcare involve physician behavior relative to individual patients, but the Medicare Part D dataset lacks patient-level information, limiting our ability to capture such patterns. As discussed below, we instead focus on scenarios observable at the physician-claim level, specifically, price-based prescribing preferences and opioid-related rules.

\begin{figure}[t]
\centering

\subfloat[]{
  \includegraphics[width=0.48\linewidth, height=33mm]{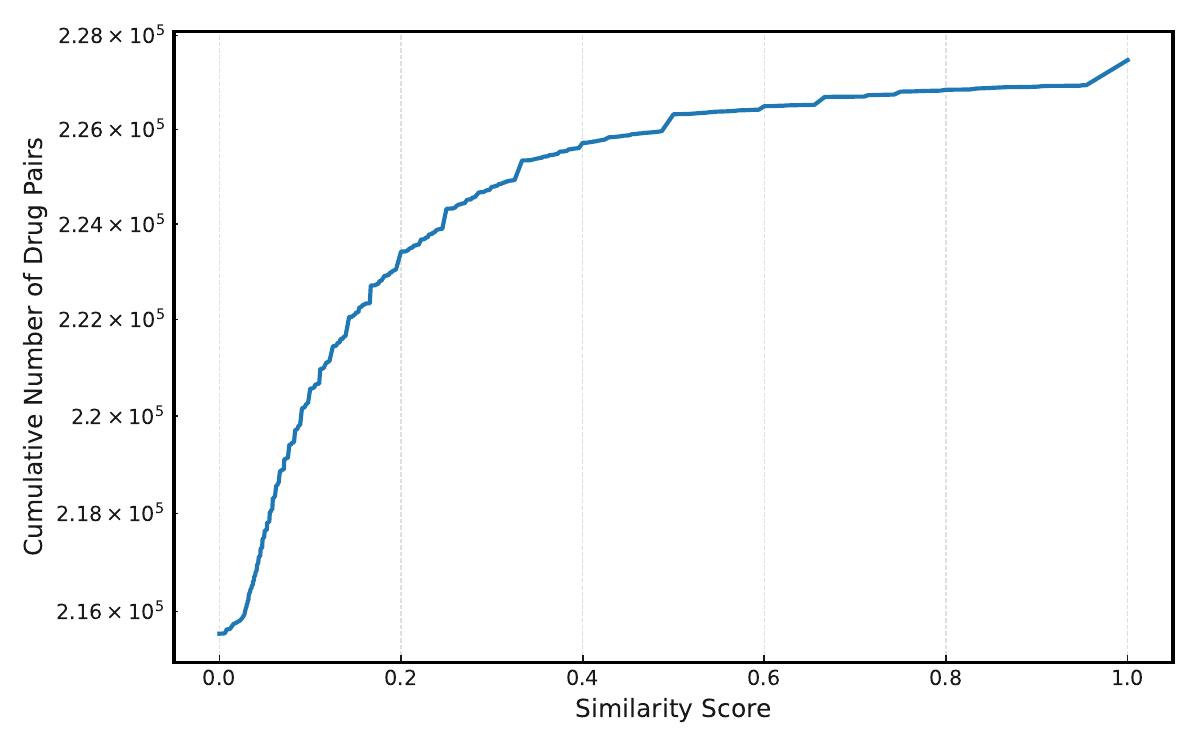}
  \label{fig:drug-similarity}
}
\subfloat[]{
  \includegraphics[width=0.48\linewidth, height=33mm]{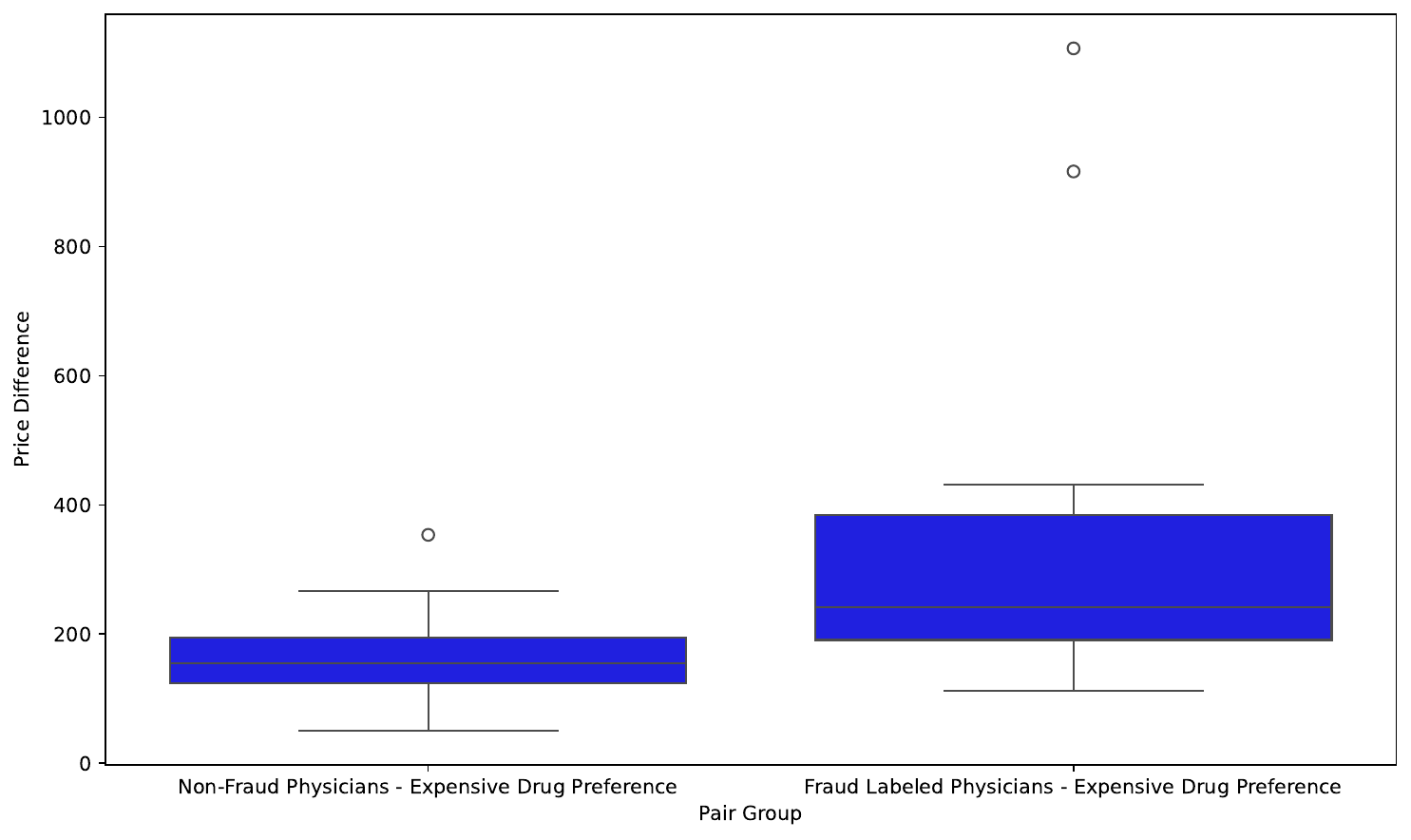}
  \label{fig:drug-price-boxplot}
}

\vspace{2mm} 

\subfloat[]{
  \includegraphics[width=\linewidth, height=33mm]{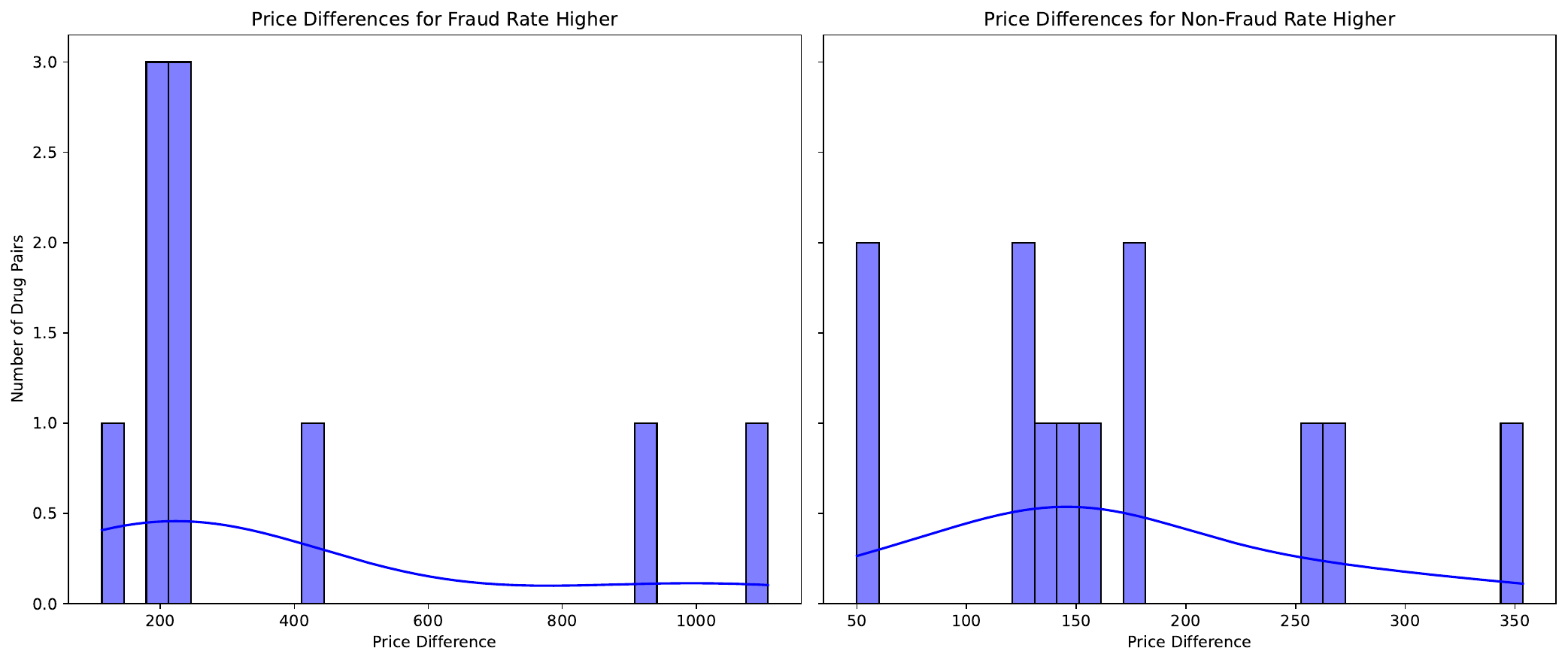}
  \label{fig:drug-price-histogram}
}

\caption{(a) Cumulative distribution of drug pair similarities by shared protein targets; (b) Box plot of price differences, grouped by whether fraud or non-fraud physicians more often prescribed the costlier drug; (c) Distribution of price differences, grouped as in (b).}

\label{fig:price}
\end{figure}
\vspace{2mm}
\subsubsection{Preference for Expensive Equivalent Drugs} 
\label{sec:preference}
To examine whether physicians preferentially prescribe higher-cost drugs despite the availability of clinically equivalent alternatives, we begin by identifying sets of interchangeable medications. Drug similarity is assessed by comparing protein target sets, a method well established in the literature~\cite{cheng2019network}. The rationale is that drugs sharing protein targets often exhibit similar mechanisms of action and therapeutic use. It is important to emphasize that, although target-identical drugs are mechanistically and often therapeutically similar, we use the term {\it interchangeable medications} strictly as an analytical construct for pricing analyses. This definition does not imply FDA-designated therapeutic equivalence or biosimilar interchangeability.


Prior work substantiates this connection. Keiser et al.\cite{Keiser2009} demonstrated that mapping drugs to protein targets via the {\it Similarity Ensemble Approach} not only recovers known pharmacological relationships but also predicts novel drug–target links with experimental validation, thereby supporting therapeutic similarity through shared targets. Wang et al.\cite{Wang2013} showed that integrating drug–target networks enhances the prediction of therapeutic classes, reinforcing the alignment between target overlap and clinical use. Complementing these perspectives, Campillos et al.~\cite{Campillos2008} inferred target similarity from clinical side-effect profiles across hundreds of marketed drugs, highlighting how target overlap influences patient-level phenotypes and suggesting new therapeutic applications.

Building on these insights, we operationalize drug similarity by computing the Jaccard index between each drug pair’s set of protein targets, that is, the ratio of the intersection to the union. Protein target sets were curated from DrugBank~\cite{knox2024drugbank}, providing a foundation for defining groups of interchangeable medications. These equivalence classes represent clinically substitutable options and allow us to focus on prescribing decisions where alternatives exist. Within each group, we then compare patterns across cost tiers to identify cases where higher-cost drugs are systematically preferred despite the availability of lower-cost, therapeutically comparable alternatives.


Figure~\ref{fig:drug-similarity} shows the cumulative number of drug pairs by similarity level. Over 95\% share no common targets, with the curve rising sharply at zero and tapering off as similarity increases, highlighting the rarity of highly similar pairs. We retained only clusters where all drugs had a Jaccard similarity of 1, indicating identical molecular targets and enabling target-based interchangeability.


Within these high-similarity clusters, we aggregated prescription data at the drug level to compute total cost, total claims, and average cost per claim (total cost divided by claims). Drugs in each cluster were then compared pairwise to assess pricing disparities. Based on the magnitude of observed gaps, pairs were categorized as moderate, high, or extreme cost-difference cases. This filtering produced 376 unique pairs, of which 245 fell into the extreme category.


Analysis of physician prescribing behavior revealed a consistent pattern. Among physicians who prescribed both drugs in a pair, those flagged as fraudulent more frequently favored the higher-cost option, particularly in cases with extreme price differences. In pairs where the preference gap exceeded 20 percentage points, the associated cost disparities were especially pronounced. These findings are reinforced by visual patterns in the data. Figure~\ref{fig:drug-price-boxplot} shows that drug pairs preferentially prescribed by fraud-labeled physicians exhibit larger median price gaps and greater variability. Similarly, Figure~\ref{fig:drug-price-histogram} highlights a broad, right-skewed price distribution with multiple peaks and outliers, in contrast to the narrower, more uniform distribution observed among non-fraud physicians.


 Based on these observations, we define a rule to capture potential fraud: preferential prescribing of a higher-cost drug within an interchangeable pair. For each physician prescribing both drugs in a pair, we compare total claims and related features, flagging those who consistently favor the costlier alternative. Multiple independent sources validate this heuristic as a meaningful fraud, waste, and abuse signal rather than random noise. For example, a New York State Comptroller audit documented over \$1.1 million in overpayments where brand-name drugs were reimbursed despite generic availability and without “dispense as written” codes, directly linking brand–generic price gaps to billing and control failures~\cite{OSC2022BrandDrugs}. Moreover, legal analysis in U.S. Pharmacist further explains that pharmacies are required to bill payers at their usual and customary price, and that charging above widely available discount rates can constitute false claims \cite{Dowell2023FCA}.
 

\subsubsection{Detection in Opioid Prescriptions}
\label{sec:opiod}

We constructed opioid-related prescription preference rules based on clinical judgment and insights from prior research on opioid misuse and overprescribing trends. Previous studies have shown that inappropriate reliance on opioids, particularly in situations where safer or equally effective non-opioid alternatives are available, can serve as a signal of problematic or even fraudulent prescribing behavior \cite{zafari2019topic, guy2017vitalsigns}. Overprescribing patterns have been associated not only with increased risk of dependence and overdose but also with deliberate exploitation of insurance reimbursement systems. Drawing on these findings, we designed rules to flag cases where opioid prescriptions occur at unusually high rates relative to established clinical norms \cite{CMS2017Opioids}.  

To operationalize these rules, we compiled a structured dataset of generic opioid drugs, each annotated with a fraud likelihood label (either low or high), derived from medical literature and expert heuristics \cite{zafari2019topic, guy2017vitalsigns}. This Medicare Part D dataset includes 2101 unique drugs, of which 37 were classified as having a high likelihood of being associated with fraudulent prescribing. These annotations informed the construction of unary rules and associated scoring thresholds. We applied a methodology consistent with that of our cost-preference framework, using domain-informed drug mappings to isolate suspicious prescription patterns in physician-level data.

\section{Preliminaries and Problem Setup}\label{sec:prelim}

\subsection{Notation}
Let $\mathcal{X}\subset\mathbb{R}^D$ denote the data space, where each sample $x\in\mathcal{X}$ is represented by a $D$–dimensional feature vector. Then, $x[p]$ denotes the $p$-th feature of $x$. Let $\mathcal{R}$ be our rule set, partitioned into
\[
\begin{aligned}
\mathcal{R}_1 &= \{(p,q,w)\mid p,q\in\{1,\dots,D\},\ p\neq q,\ w\in[0,1]\}\\
\mathcal{R}_2 &= \{(p,w)\mid p\in\{1,\dots,D\},\ w\in[0,1]\}
\end{aligned}
\]

\noindent where each $(p,q,w)\in\mathcal{R}_1$ is a \textit{binary} rule (“feature $p$ should exceed feature $q$”) with an associated weight $w$, and each $(p,w)\in\mathcal{R}_2$ is a \textit{unary} rule (“feature $p$ should be high”) with weight $w$. 

We let $\varrho:\mathcal{R}\to\mathbb{R}^L$ be a pretrained \textit{Rule Encoder} \texttt{(RE)} that maps each rule to an $L$-dimensional embedding, and $\phi:\mathbb{R}^D\to\mathbb{R}^L$ be a \textit{Sample Encoder} \texttt{(SE)} mapping data samples into the same latent space. A {\it base anomaly detection model} $f_{\theta_1}: \mathbb{R}^D \rightarrow [0,1]$ assigns a score to each data sample relevant to its probability of being an anomaly; this model is hereafter referred to as {\texttt{BASE}. Denote by $d(u,v)=\|u-v\|_2$ the Euclidean distance in $\mathbb{R}^L$.

The {\texttt{BASE} loss function is denoted by $\mathcal{L}_{\texttt{BASE}}(f_{\theta_1}(\mathcal{X}), \mathcal{Y})$, which captures semi-supervised performance on the labeled subset of the dataset. The {\it alignment loss} is defined as $\mathcal{L}_{\text{align}}(\phi_{\theta_3}(\mathcal{S}), \varrho_{\theta_2}(\mathcal{R}))$, and quantifies how well data samples $\mathcal{S}\subset \mathbb{R}^D$ conforms to the set of domain rules. Minimizing this loss yields a soft alignment score for every input $x \in \mathcal{X}$, denoted by $\mathcal{C}(x, \mathcal{R})$, which reflects the degree to which $x$ satisfies the rule set. Table~\ref{tab:notation} shows a full summary of notations.

\subsection{Knowledge Rules for Fraud Detection}
We encode domain knowledge about anomalous prescribing behavior as weighted rules:

\begin{itemize}[leftmargin=*, label=$\diamond$]
  \item \textbf{Cost‐preference rules} $(p,q,w)$: For two equivalent drugs indexed by $p$ and $q$, we expect $x[p]$ (claims for the higher‐cost drug) to exceed $x[q]$ only to a modest degree. A large violation (high $x[p]-x[q]$) with high weight $w$ indicates potential fraud.
  \vspace{2mm}
  \item \textbf{Opioid‐Prescription rules} $(p,w)$: For feature $p$ encoding a physician’s aggregate opioid‐topic score, we expect $x[p]$ to lie below a predefined threshold. A value exceeding this threshold indicates suspicious prescribing behavior and incurs a penalty proportional to $w$. 
\end{itemize}

These rules are applied to relevant features such as total cost, claim count, beneficiaries, 30-day fill count, and days of supply. Each rule is weighted by confidence $w \in [0,1]$. For cost-preference rules, the weight is calculated based on the price difference between the more expensive drug and its less expensive equivalent. For opioid prescription rules, the weight reflects the likelihood that prescribing a given drug is associated with fraudulent behavior. We refer to these domain-informed guidelines as \textit{rules} hereafter.


\subsection{Problem Statement}

Given a dataset of provider feature vectors $\mathcal{X} = \{x_i\}_{i=1}^N$ where $x_i \in \mathbb{R}^D$, we assume access to:

\begin{itemize}[leftmargin=*, label=$\diamond$]
   \item A small labeled subset $\mathcal{X}_L$ with binary fraud indicators $\mathcal{Y}$.
  \item A large unlabeled subset $\mathcal{X}_U = \mathcal{X} \setminus \mathcal{X}_L$.
  \item A set of soft, expert-defined rules $\mathcal{R}$ describing suspicious prescribing behavior.
\end{itemize}

Our goal is to learn an {\it anomaly scoring} function $f: \mathbb{R}^D \to [0,1]$ that ranks fraudulent physicians above benign ones. The model should leverage both labeled data and the structured knowledge encoded in $\mathcal{R}$. To achieve this, we later define a composite learning objective that combines \{supervised} loss with a rule-alignment component.

\begin{table}[t!]
\centering
\caption{Summary of Notation}
\label{tab:notation}
\begin{tabular}{@{}ll@{}}
\toprule
\textbf{Symbol} & \textbf{Description} \\
\midrule
$\mathcal{X}$ & Input data space, $\mathcal{X} \subset \mathbb{R}^D$ \\
$\mathcal{X}_L$ & Labeled subset of data \\
$\mathcal{X}_U$ & Unlabeled subset of data \\
$x$ & A single data sample $x \in \mathbb{R}^D$ \\
$x[p]$ & $p$-th feature of sample $x$ \\
$\Delta_x$ & rule-contrast feature vector of $x$\\
$D$ & Number of input features \\
$\mathcal{R}$ & Set of symbolic rules \\
$\mathcal{R}_1$ & Set of binary rules: $(p,q,w)$ \\
$\mathcal{R}_2$ & Set of unary rules: $(p,w)$ \\
$w$ & Rule confidence weight in $[0,1]$ \\
$\delta(r, x)$ & 1 if rule $r$ applies to sample $x$, else 0 \\
$\varrho(\cdot)$ & Rule Encoder: $\mathcal{R} \rightarrow \mathbb{R}^k$ \\
$\phi(\cdot)$ & Sample Encoder: $\mathbb{R}^D \rightarrow \mathbb{R}^k$ \\
$e_r$ & Rule embedding $\varrho(r)$ \\
$e_x$ & Sample embedding $\phi(x)$ \\
$\text{sim}(u,v)$ & Similarity function (e.g., cosine similarity) \\
$d(u,v)$ & Euclidean distance: $\|u - v\|_2$ \\
$\mathcal{L}_{\text{task}}$ & Base model loss \\
$\mathcal{L}_{\text{triplet}}$ & Triplet loss aligning rule/sample embeddings \\
$\mathcal{L}_{\text{OT}}$ & Optimal Transport alignment loss \\
$\lambda_1, \lambda_2$ & Hyperparameters for weighting losses \\
$f(x)$ & Final scoring function over sample embeddings \\
\bottomrule
\end{tabular}
\end{table}

\vspace{2mm}

\section{Methodology}\label{sec:methodology}
\cc~ incorporates structured domain knowledge into a base fraud detection model (\texttt{BASE}), especially effective in high-dimensional settings with limited labeled data. Knowledge is encoded as simple, interpretable pairwise rules (e.g., “feature $p$ should be high and $q$ low”), each weighted by a confidence score. Instead of applying rules in the input space, both rules and data samples are embedded into a shared low-dimensional space $\mathbb{R}^L$ using two neural networks: \texttt{RE} for rules and \texttt{SE} for samples. Alignment is measured via Euclidean distance: samples satisfying a rule are embedded closer to it, whereas violators lie farther away. This geometric alignment regularizes the anomaly detector, allowing it to integrate domain knowledge in a flexible way.
Since domain rules capture general fraud patterns and are dataset-independent, \texttt{RE} and \texttt{SE} are trained on synthetic data. This data consists of artificially generated samples that explicitly satisfy or violate individual rules. For each rule, a positive (satisfying) and negative (violating) sample are generated, and a weighted triplet loss is used to train both encoders. To ensure stable convergence, training alternates between updating one encoder while keeping the other fixed, encouraging embeddings where rules lie closer to satisfying than violating samples. Each rule is assigned a confidence weight that serves two purposes: guiding the sampling toward more reliable rules and scaling the triplet loss to increase their impact on the embedding space.

By training solely on synthetic data, \cc~ avoids dataset-specific biases and learns soft, continuous rule representations in a latent space. This enables flexible, interpretable fraud detection without hard constraints, supporting generalizable reasoning on unseen data. In the following, we first describe the general framework of \cc~ and next, we show how we can apply this framework to our fraud detection problem.

\subsection{Rule-Contrast Feature Engineering}
\label{sec:featureEngineering}

Raw prescription totals are high-dimensional and heavily skewed by prescriber volume, making them unsuitable for direct comparison across physicians or over time. To enable meaningful and compact representations, we transform the data into rule-aligned features that capture relative prescribing tendencies rather than absolute counts. This feature engineering step provides a standardized way to highlight potentially concerning behaviors as defined by the rules introduced in Sections~\ref{sec:preference} and~\ref{sec:opiod}.

To normalize magnitudes across drugs and prescribers, we convert each raw total into a \textit{Prescriber-Year Share} (bounded in $[0,1]$).
For a metric $m\in\{\text{Clm},\text{Fill30},\text{Days},\text{Cost},\text{Bene}\}$, define the denominator
\[
S^{(m)}_{t,i}=\sum_{d}\text{Tot}^{(m)}_{t,i,d},
\]
and the Prescriber-Year Share
\[
\text{share}^{(m)}_{t,i,d} \;=\;
\begin{cases}
\text{Tot}^{(m)}_{t,i,d}\big/ S^{(m)}_{t,i}, & S^{(m)}_{t,i}>0,\\[2pt]
0, & S^{(m)}_{t,i}=0,
\end{cases}
\]
so that $\sum_d \text{share}^{(m)}_{t,i,d}=1$ whenever $S^{(m)}_{t,i}>0$.  
Intuitively, the Prescriber-Year Share represents the fraction of prescriber $i$’s total prescribing activity in year $t$ that is devoted to drug $d$ under metric $m$.
For example, if prescriber $i$ issued 100 total claims in year $t$, of which 20 were for drug $d$, then $\text{share}^{(\text{Clm})}_{t,i,d}=0.2$.

Next, for each rule $j=(p_j,q_j,w_j)$, year $t$, and prescriber $i$, we compute five \textit{rule-contrast coordinates}, one for each metric channel:
\[
\Delta^{(m)}_{i,t}(j) = \text{share}^{(m)}_{t,i,p_j} - \text{share}^{(m)}_{t,i,q_j}, 
\]
\[
m \in \{\text{Clm}, \text{Fill30}, \text{Days}, \text{Cost}, \text{Bene}\}.
\]
A positive contrast indicates that the drug {\it more concerning} $p_j$ constitutes a larger share than its comparator $q_j$ under metric $m$.


To capture longitudinal behavior, we aggregate each contrast across years using three statistical summaries:
\[
\begin{aligned}
\Phi^{(m)}_{\min,i}(j) \;&=\; \min_{t\in\mathcal{T}} \Delta^{(m)}_{i,t}(j), \\[4pt]
\Phi^{(m)}_{\text{mean},i}(j) \;&=\; \tfrac{1}{|\mathcal{T}|}\sum_{t\in\mathcal{T}} \Delta^{(m)}_{i,t}(j), \\[4pt]
\Phi^{(m)}_{\max,i}(j) \;&=\; \max_{t\in\mathcal{T}} \Delta^{(m)}_{i,t}(j).
\end{aligned}
\]

The final feature vector for prescriber $i$ is obtained by concatenating, over all rules $j$ and all five channels $m$, the three temporal summaries above. This yields a compact, rule-aligned representation of dimension $15R$ (five channels $\times$ three aggregations per rule).

\begin{figure}[t]
  \centering
  \includegraphics [width=\linewidth]{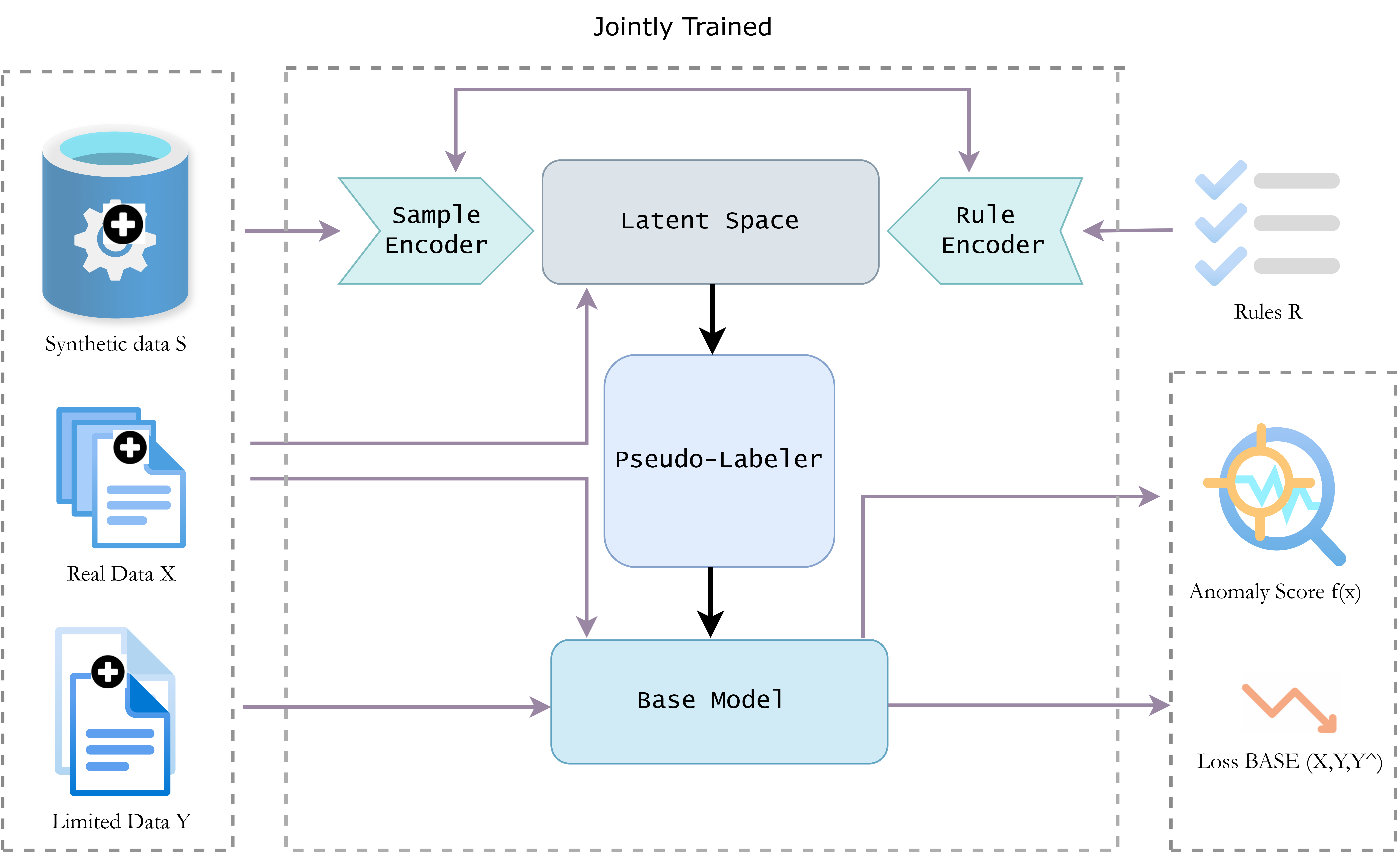}

  \caption{Workflow of \cc. The system embeds domain rules and prescription data into a shared latent space via rule and sample encoders. Synthetic compliance/violation samples guide alignment, which generates pseudo-labels to enhance the base anomaly detection model.}
  \label{fig: Workflow  }
\end{figure}

\subsection{Rule and Data Embedding}
\noindent \textbf{Rule Encoder (RE).}
The \texttt{RE} maps each pairwise fraud detection rule into a vector in the latent embedding space $\mathbb{R}^L$. Each binary rule is defined by a tuple $(p, q)$, where $p, q \in \{1, \dots, D\}$ index features in the original input space. The encoder first embeds each individual feature index into a lower-dimensional intermediate space via a learnable embedding matrix $\mathbf{E} \in \mathbb{R}^{D \times d}$, where $d \ll D$. The embedding vectors corresponding to $p$ and $q$ (i.e., $\mathbf{e}_p$ and $\mathbf{e}_q$) are then concatenated into a single vector $[\mathbf{e}_p; \mathbf{e}_q] \in \mathbb{R}^{2d}$. This concatenated representation is passed through a multilayer perceptron (MLP), which transforms it into a rule embedding $\varrho_{\theta_2}(p, q) \in \mathbb{R}^L$ in the shared latent space. Formally, the RE can be expressed as:
\[
\varrho_{\theta_2}(p, q) = \text{MLP}\left([\mathbf{E}_p; \mathbf{E}_q]\right)
\]

To handle unary rules (e.g., “feature \(p\) should be high/low”), a special learnable vector \(\mathbf{e}_{\text{NULL}}\) acts as a placeholder. Unary rules are encoded as \(\varrho(p, \text{NULL}) = \text{MLP}([\mathbf{e}_p; \mathbf{e}_{\text{NULL}}])\), allowing both unary and binary rules to share the same encoding framework.

This approach is both scalable and expressive. By operating directly on feature indices, the model avoids the need for manually encoding rule logic, while jointly learning an embedding matrix across all rules captures shared semantics among feature positions. The use of a shared embedding space also enables the model to generalize to unseen rule pairs and distinguish between similar structures (e.g., $(p, q)$ vs. $(q, p)$). \\

\noindent \textbf{Sample Encoder (SE).} The \texttt{SE} maps data samples to the same latent space $\mathbb{R}^L$ as the \texttt{RE}, enabling distance-based reasoning over the satisfaction of the rules. Suppose that our rules set $\mathcal{R} = \{ (p_j, q_j, w_j) \}_{j=1}^{R},
$ where $p_j$ and $q_j$ are features of data samples (for unary rules, we assume $q_j=e$, a feature with zero value), with $w_j \in [0,1]$ encoding the importance of the rules (e.g. cost or severity of opioids). For each data sample $x$, the \textit{rule-contrast feature vector} $\Delta_x \in \mathbb{R}^{15R}$ is constructed as in \ref{sec:featureEngineering}. These vectors are then passed to the \texttt{SE} for comparison with rule embeddings. The reason we work with rule-contrast feature vectors instead of directly passing the original data samples $x$ to \texttt{SE} is that usually $R\ll D$, which means reducing the dimension of the \texttt{SE} input improves trainability and stability of the trained network.

Note that \texttt{SE} is trained on synthetic examples specifically constructed to satisfy or violate known rules. These synthetic data consist of vectors $\Delta\subset \mathbb{R}^{15R}$ such that high values of elements corresponding to their $j^{th}$-rule means satisfying the $j^{th}$-rule. These synthetic rows are generated based on the weights of the rules in $\mathcal{R}$ (a rule with higher weight will appear more in the synthetic data and thus the network will be more sensitive to such rules). For each data sample $x$, \texttt{SE} outputs a latent embedding $\phi_{\theta_3}(x)$, implemented as a lightweight MLP on top of $\Delta_x$. During pre-training, \texttt{SE} and \texttt{RE} are jointly optimized using a weighted triplet loss, which encourages satisfying samples to be embedded closer to their corresponding rule vectors than violating ones. This training strategy allows the model to learn a geometry that reflects the compliance of the rules in a continuous and generalizable way.


\vspace{2mm}


\noindent \textbf{Weighted Triplet Loss.} 
To train the \texttt{RE} and \texttt{SE} jointly, we employ a weighted triplet loss that encourages alignment between rule embeddings and data samples that satisfy those rules. For each rule $r_j$, we use synthetic examples, previously described, to construct satisfying (positive) and violating (negative) samples. Let $\varrho(r_j) \in \mathbb{R}^L$ denote the rule embedding, and $\phi(x^+), \phi(x^-)$ be the embeddings of the positive and negative samples, respectively. The triplet loss aims to enforce:
\[
\|\phi(x^+) - \varrho(r_j)\|_2^2 + \text{margin} < \|\phi(x^-) - \varrho(r_j)\|_2^2
\]
by minimizing the hinge-based objective:


\[
\mathcal{L}_{\text{triplet}} = w_j \cdot \max\Biggl(0,\;
\begin{aligned}
   & \|\phi(x^+) - \varrho(r_j)\|_2^2 \\
   & - \|\phi(x^-) - \varrho(r_j)\|_2^2 + m
\end{aligned}
\Biggr)
\]
where $w_j$ is the confidence weight assigned to rule $r_j$, and $m$ is a predefined margin. This weighted formulation ensures that high-confidence rules have a greater impact on the embedding space, guiding the encoders to prioritize rules that reflect more reliable or impactful domain knowledge. 

\vspace{2mm}
\noindent \textbf{Augmenting the Base Model with Domain Knowledge.}  
To leverage domain knowledge
in the absence of extensive supervision, we generate pseudo-labels for each data sample based on its alignment with the set of encoded rules. For each mini-batch $\mathcal{B} = \{x_i\}_{i=1}^B \subset \mathcal{X}$, we compute pseudo-labels using optimal transport (OT) alignment between the embedded samples and the rule set. Let $\phi_{\theta_3}(\mathcal{B}) \in \mathbb{R}^{B \times L}$ be the latent representations of the batch, and let $\varrho_{\theta_2}(\mathcal{R}) = \{r_j\}_{j=1}^R \subset \mathbb{R}^L$ be the set of encoded rules. We define the OT transport plan $T \in \mathbb{R}^{B \times R}$ between the batch and rule embeddings via Sinkhorn iterations \cite{knight2008sinkhorn}, using a regularized kernel. From this plan, we compute a per-sample transport cost:

\[
c_i = \frac{\sum_{j=1}^R T_{ij}\,\|\phi_{\theta_3}(x_i)-r_j\|_2^2}{\sum_{j=1}^R T_{ij}}
\]
yielding pseudo-labels,
\[
\hat y_i = \sigma\!\left(\frac{\mu - c_i}{\tau\,s+\varepsilon}\right)
\]
\noindent with $\mu,s$ global running mean/stdev,
$\tau>0$ is a fixed \textit{temperature} controlling the sharpness of the sigmoid mapping and $\epsilon>0$ is a constant to increase stability. Here, $\hat{y}_i \in [0, 1]$ reflects how well sample $x_i$ aligns with the domain rules. We then integrate these pseudo-labels into the training of the \texttt{BASE} model $f_{\theta_1}: \mathbb{R}^D \to [0,1]$, using a hybrid objective:
\vspace{1mm}


\[
\begin{aligned}
\mathcal{L}_{\text{total}} = &\; 
   \frac{1}{|\mathcal{L}|} \sum_{i \in \mathcal{L}} 
   \mathcal{L}_{\texttt{BASE}}(f_{\theta_1}(x_i), y_i) \\
   & + \lambda \cdot \frac{1}{B} \sum_{i \in \mathcal{B}} 
   \mathcal{L}_{\text{align}}(f_{\theta_1}(x_i), \hat{y}_i)
\end{aligned}
\]

\noindent where $\lambda$ is the confidence factor.

\section{Experiments}\label{sec:experiments}
In this section, we present experimental results to evaluate the effectiveness of \cc, in comparison to baselines for weakly supervised anomaly detection. We aim to answer two key research questions. First, how effective is \cc, which integrates structured domain knowledge, compared to baselines? Second, what is the contribution of the knowledge-data alignment mechanism to the overall performance of \cc? These guide our analysis of the value of incorporating soft domain rules in fraud detection.

\subsection{Experimental Design}
\noindent {\bf Dataset.} For our experiments, we use the Medicare Part D dataset introduced in Section~\ref{sec:dataCharacteristics}, which, to the best of our knowledge, is the only publicly available dataset for healthcare fraud detection. Building on the domain-informed rules defined earlier, we extract 376 binary rules from physician prescribing preferences (Section~\ref{sec:preference}) and 37 unary rules from opioid-related prescribing patterns (Section~\ref{sec:opiod}). Together, these yield a total of $R=413$ rule-based pairs. Applying the rule-contrast feature reduction technique compresses the original 2{,}101 drug features to 413, resulting in a more compact and structured dataset for analysis.

\vspace{2mm}

\noindent {\bf Evaluation Metrics.} We evaluate all methods using two widely adopted metrics: AUC and standard classification metrics, including precision, recall, and F1 score, computed at a fixed threshold. AUC refers to the area under the Precision-Recall curve across varying thresholds. Precision measures the proportion of true anomalies among the instances flagged by the model. Recall measures the proportion of true anomalies that were correctly identified. F1 score captures the balance between precision and recall.

\vspace{2mm}

\noindent {\bf Baselines.} In the absence of weakly supervised learning methods specifically tailored for fraud detection, we compare our approach with the following baselines for anomaly detection:

\begin{itemize}[leftmargin=*, label=$\diamond$]
    \item {\pt MLP} \cite{rumelhart1986learning}: A Multi-Layer Perceptron (MLP) is a neural network with fully connected layers for learning non-linear patterns.
    \item {\pt DeepSAD} \cite{ruff2020deepsemisupervisedanomalydetection}: A deep semi-supervised one-class classification method that enhances an unsupervised framework.
    \item {\pt DevNet} \cite{pang2019deep}: A neural network-based model trained using a deviation loss function to identify anomalies.
    \item {\pt PReNet} \cite{pang2023deep}: A neural network-based model that employs a two-stream ordinal regression approach to learn relationships between instance pairs.
\end{itemize}

\vspace{2mm}

\noindent{\bf Model Variants for Evaluation.} To evaluate the effectiveness of our rule-informed learning framework for fraud detection, we use two versions of \texttt{BASE} models:

\begin{itemize}[leftmargin=*, label=$\diamond$]
    \item \noindent\textbf{Baseline Evaluation:} Assess the standalone performance of each \texttt{BASE} model using standard fraud detection metrics to establish a reference point. This version of the baseline model will be referred to by its name only. 
   
    
   
    \item \noindent\textbf{Knowledge Alignment:} Integrate alignment scores into each \texttt{BASE} model via our knowledge injection techniques and measure the resulting performance improvements across all models. We refer to this version of the baseline model as \cc-Baseline.
\end{itemize}

\subsection{Comparative Evaluation}
The performance metrics presented in Table~\ref{tab:model-performance} illustrate the impact of incorporating domain knowledge through \cc~ in baseline models. Evaluated using both AUC and Recall at Top-K thresholds (R@K), the results show that embedding structured domain semantics can significantly enhance the models' ability to prioritize fraudulent instances.

Among the baseline models, MLP achieves the highest AUC (0.84), followed by DevNet (0.79), PReNet (0.74), and DeepSAD (0.71). When enhanced with \texttt{CleverCatch}, most models show improvements in recall metrics, and in several cases, modest gains in AUC. For example, \texttt{CleverCatch-DeepSAD} improves AUC from 0.71 to 0.73 and substantially boosts R@100 from 0.182 to 0.215, indicating a much stronger ability to retrieve relevant anomalies in the top-ranked predictions. Similarly, \texttt{CleverCatch-DevNet} and \texttt{CleverCatch-PReNet} both improve recall at every threshold compared to their baseline counterparts, while maintaining competitive AUC values (0.80 and 0.75, respectively).

Even in cases where AUC remains stable or slightly decreases—as with MLP (from 0.84 to 0.83)—\cc~ provides a notable improvement in recall, e.g., increasing R@100 from 0.038 to 0.045. This suggests that domain-guided alignment enables the model to prioritize the most critical cases better, even if the overall discrimination boundary does not shift substantially.

These findings collectively indicate that the integration of domain knowledge via \cc~ complements data-driven learning. Rather than relying solely on statistical patterns, models benefit from structured domain rules that guide the alignment of feature representations toward semantically meaningful distinctions. This results in more effective anomaly ranking across diverse architectures and thresholds, demonstrating the general effectiveness of the proposed approach.

\subsection{Ablation Study}
Between the two domain rule types, cost-preference rules contribute more significantly to performance gains than opioid-drug rules. On average, removing cost-preference rules leads to a drop $21\%$ in R@100 across the four evaluated models. In comparison, removing opioid-related rules results in a smaller performance decline $8\%$ in R@100. Although their impact differs, the rules are complementary: cost-preference rules primarily capture economically motivated anomalies, while opioid rules identify clinical misuse patterns. Together, they enable \cc\ to detect a broader spectrum of fraudulent behaviors overlooked by purely data-driven models. If supported by training data, these rules can be extended to capture emerging or domain-specific fraud patterns.

From our experiments, we observed that if we use only pseudo-labels, generated independently of the \texttt{BASE} model, to classify the data by assigning a label of 1 to any instance with a pseudo-label above the 50\% threshold, we achieve an AUC score of approximately 0.64. This suggests that our rule-based alignment model is indeed sensitive to patterns characteristic of fraudulent NPIs. Furthermore, when comparing pseudo-label predictions with those of the \texttt{BASE} model, we find that while there is agreement on clear-cut cases, the predictions are not entirely overlapping. This indicates that the pseudo-labels and \texttt{BASE} models capture complementary aspects of fraud behavior, reinforcing the value of integrating structured domain knowledge into the learning process.

These findings reinforce the value of integrating domain-specific heuristics into machine learning frameworks for fraud detection. By capturing statistical irregularities and normative violations, \cc\ leverages domain knowledge to bridge the gap between empirical patterns and expert-driven expectations.

\begin{table}[t]
\centering
\caption{Performance comparison of the baseline and its variants with respect to AUC and R@K at different thresholds}
\begin{tabular}{l c cccc}
\toprule
\textbf{Model} & \textbf{AUC} & \multicolumn{4}{c}{\textbf{R@K}} \\
\cmidrule(lr){3-6}
& & \textbf{@10} & \textbf{@20} & \textbf{@50} & \textbf{@100} \\
\midrule
\pt MLP              & 0.84 & 0.004 & 0.008 & 0.019 & 0.038 \\
\pt CleverCatch-MLP  & 0.83 & 0.005 & 0.010 & 0.023 & 0.045 \\
\midrule
\pt DeepSAD              & 0.71 & 0.018 & 0.036 & 0.099 & 0.182 \\
\pt CleverCatch-DeepSAD  & 0.73 & 0.020 & 0.040 & 0.115 & 0.215 \\
\midrule
\pt DevNet              & 0.79 & 0.011 & 0.027 & 0.066 & 0.127 \\
\pt CleverCatch-DevNet  & 0.80 & 0.012 & 0.031 & 0.081 & 0.155 \\
\midrule
\pt PReNet              & 0.74 & 0.010 & 0.023 & 0.055 & 0.105 \\
\pt CleverCatch-PReNet  & 0.75 & 0.011 & 0.027 & 0.066 & 0.132 \\
\bottomrule
\end{tabular}
\label{tab:model-performance}
\end{table}

}

\section{Related Work}\label{sec:related}

\subsection {Conventional Fraud Detection} 
Supervised models approach healthcare fraud detection as a binary classification task. Bauder et al.~\cite{bauder2016outlier} used Naive Bayes to flag physicians submitting atypical claims. Later work introduced cost estimation to detect anomalies, with multivariate adaptive regression splines performing best. Unsupervised models identify outliers without labeled data. Herland et al. \cite{sadiq2017primm} detect fraud by locating high-density anomalous regions. Suesserman et al. used unsupervised autoencoders with a feature-weighted loss to detect procedure overutilization in healthcare claims, showing strong results without labeled data \cite{suesserman2023procedure}. Johnson and Khoshgoftaar~\cite{johnson2019medicare} found that over-aggressive downsampling harms class-imbalanced fraud detection. Deep learning models offer strong capabilities for fraud detection via representation learning or anomaly scoring~\cite{pang2021deep}. While their use in healthcare is still limited, data fusion, especially between Medicare sources, has been shown to be critical to improving performance~\cite{herland2018bigdata}.

Recent trends, as explained in the following sections, also indicate the emergence of hybrid strategies that combine different techniques within ensemble frameworks. Such methods aim to exploit the complementary strengths of each paradigm: the interpretability of rule-based or supervised models, the adaptability of unsupervised anomaly detection, and the pattern recognition power of deep neural networks. As healthcare fraud schemes evolve in sophistication, these multi-layered approaches are increasingly viewed as essential for scalable, real-world fraud detection systems.


\subsection{Weak Supervision Models} 
Fraud detection is challenged by scarce anomaly labels, noisy data, and evolving patterns, making fully supervised learning impractical. While common in anomaly detection, weak supervision is underused in fraud detection, but it can be effectively adapted for it. Weakly supervised anomaly detection addresses this by leveraging limited or imperfect labels. DeepSAD \cite{ruff2020deepsemisupervisedanomalydetection} builds on Deep SVDD \cite{pmlr-v80-ruff18a}, using labeled normal and anomalous data to separate them in the latent space. It performs well even with limited labels. DevNet \cite{pang2019deep} targets extreme label sparsity using a Gaussian prior and deviation-based loss, producing interpretable scores validated on real-world fraud data. PReNet \cite{pang2023deep} uses pairwise comparisons to learn discriminative features, enabling detection of both known and novel anomalies.

From a methodological perspective, these approaches present different strategies for integrating weak supervision into deep anomaly detection: embedding guidance (DeepSAD), distributional regularization (DevNet), and relative learning (PReNet). Their success suggests that fraud detection systems can benefit from hybrid pipelines, where domain-informed heuristics provide weak labels that are refined by deep models. This not only mitigates the scarcity of high-quality fraud annotations but also yields models that remain adaptive to evolving fraudulent behaviors.

\subsection{Knowledge-Guided Models} 
These models integrate domain expertise, such as heuristic rules, symbolic reasoning, or relational structures, into machine learning models. In healthcare fraud detection, where labeled data is scarce and fraud evolves, this approach enhances both accuracy and interpretability. Recent methods embed domain knowledge in various forms. Rao et al. \cite{rao2021knowgnn} extended this by incorporating graph functional dependencies for interpretable predictions. Symbolic methods are also used. KnowGraph~\cite{zhou2024knowgraph} integrates weighted first-order logic into GNNs, while Deep Symbolic Classification~\cite{visbeek2023dsc} discovers analytic expressions to separate fraud from non-fraud. Interpretability can be built into model structure. SEFraud~\cite{li2024sefraud} learns masks to highlight key features and edges, aligning with real-world expectations and deployed for financial fraud detection. Pan et al.~\cite{pan2025huge} encode domain knowledge of fraud connections into a heterophily-aware unsupervised model. In customs fraud, Park et al.~\cite{park2022das} show that prototypical knowledge can be transferred across regions via domain adaptation, serving as expert supervision across borders.

{Among existing approaches, KDAlign \cite{zhao2024weakly} is one of the most closely related to our work, as it also leverages domain knowledge under weak supervision. Their method requires each domain rule to be expressed in deterministic decomposable negation normal form (d-DNNF) \cite{darwiche2002knowledge}, and utilizes a graph convolutional network (GCN) \cite{kipf2016semi} to learn rule embeddings over a predefined logical graph.
However, a key distinction lies in the input structure: while KDAlign assumes a predefined graph as a knowledge structure, our setting operates over a sequence of knowledge rules, which reflects a more natural form for many real-world applications. Representing knowledge using a graph structure becomes impractical when dealing with a large number of simple (e.g., unary or binary) rules. In such cases, the graph tends to be sparse, making the GCN-based pipeline in \cite{zhao2024weakly} not only computationally expensive but also ineffective due to the limited connectivity among nodes.
Our method, \cc, addresses these challenges directly. It provides a more scalable framework for incorporating domain knowledge, particularly in scenarios like fraud detection, and can be easily adapted for a broader range of anomaly detection tasks.}

\section{Conclusion}\label{sec:conclusion}
This paper introduced \cc, a novel framework for healthcare fraud detection that embeds expert rules into a shared latent space alongside data representations. By jointly learning from synthetic examples of rule compliance and violation, \cc enables flexible reasoning about rule adherence in high-dimensional data with limited labeled anomalies. It improves upon a baseline anomaly detection model by incorporating domain knowledge, leading to stronger performance and better generalization. The approach captures complex fraud behaviors that traditional methods often miss, demonstrating the value of combining weak supervision with structured expert knowledge in real-world fraud detection. Beyond the technical contributions, this work highlights the broader significance of knowledge-guided approaches in regulated and high-stakes domains. Fraud detection systems must balance accuracy with interpretability, transparency, and fairness in order to gain acceptance from healthcare professionals, insurers, and regulators. By embedding expert-derived heuristics into the learning process, \cc~ offers not only measurable performance gains but also an interpretable decision-making framework that can support auditing, compliance, and policy alignment. This positions \cc~ as a step toward responsible AI for healthcare, where machine learning models augment expert oversight rather than replace it. Future work will extend rule sets to richer multi-modal signals, develop adaptive rule weighting to reflect evolving fraud schemes, and explore integration with real-time monitoring pipelines.

\newpage
\normalem
\bibliographystyle{IEEEtran}
\bibliography{biblio}

@article{national2018crossing, title={Crossing the global quality chasm: improving health care worldwide}, 
author={National Academies of Sciences and Medicine and Medicine Division and Board on Global Health and Committee on Improving the Quality of Health Care Globally}, 
year={2018}, 
publisher={National Academies Press}}

@inproceedings{pang2019deep,
  title={Deep anomaly detection with deviation networks},
  author={Pang, Guansong and Shen, Chunhua and van den Hengel, Anton},
  booktitle={Proceedings of the 25th ACM SIGKDD international conference on knowledge discovery \& data mining},
  pages={353--362},
  year={2019}
}

@article{rumelhart1986learning,
  title={Learning representations by back-propagating errors},
  author={Rumelhart, David E and Hinton, Geoffrey E and Williams, Ronald J},
  journal={Nature},
  volume={323},
  number={6088},
  pages={533--536},
  year={1986},
  publisher={Nature Publishing Group}
}

@inproceedings{pang2023deep,
  title={Deep Weakly-supervised Anomaly Detection},
  author={Pang, Guansong and Shen, Chunhua and Jin, Huidong and van den Hengel, Anton},
  booktitle={Proceedings of the 29th ACM SIGKDD international conference on knowledge discovery \& data mining},
  year={2023}
}

@InProceedings{pmlr-v80-ruff18a,
  title = 	 {Deep One-Class Classification},
  author =       {Ruff, Lukas and Vandermeulen, Robert and Goernitz, Nico and Deecke, Lucas and Siddiqui, Shoaib Ahmed and Binder, Alexander and M{\"u}ller, Emmanuel and Kloft, Marius},
  booktitle = 	 {Proceedings of the 35th International Conference on Machine Learning},
  pages = 	 {4393--4402},
  year = 	 {2018},
  editor = 	 {Dy, Jennifer and Krause, Andreas},
  volume = 	 {80},
  series = 	 {Proceedings of Machine Learning Research},
  month = 	 {10--15 Jul},
  publisher =    {PMLR},
  url = 	 {https://proceedings.mlr.press/v80/ruff18a.html}
}

@misc{ruff2020deepsemisupervisedanomalydetection,
      title={Deep Semi-Supervised Anomaly Detection}, 
      author={Lukas Ruff and Robert A. Vandermeulen and Nico Görnitz and Alexander Binder and Emmanuel Müller and Klaus-Robert Müller and Marius Kloft},
      year={2020},
      eprint={1906.02694},
      archivePrefix={arXiv},
      primaryClass={cs.LG},
      url={https://arxiv.org/abs/1906.02694}, 
}

@misc{nhcaa2021fraud,
author = {{National Health Care Anti-Fraud Association}},
title = {The Challenge of Health Care Fraud},
year = {2021},
howpublished = {\url{https://www.nhcaa.org/tools-insights/about-health-care-fraud/the-challenge-of-health-care-fraud/}},
note = {Accessed: 2025-05-17}
}

@misc{transparency2021health,
author = {{Transparency International}},
title = {Global Health and Corruption},
year = {2021},
howpublished = {\url{https://www.transparency.org.uk/what-we-do/global-health-and-corruption}},
note = {Accessed: 2025-05-17}
}

@misc{clhia2023fraud,
author = {{Canadian Life and Health Insurance Association}},
title = {Healthcare Anti-Fraud},
year = {2023},
howpublished = {\url{https://www.clhia.ca/web/CLHIA_LP4W_LND_Webstation.nsf/page/10D0B370160E723B85257F03005BD980}},
note = {Accessed: 2025-05-17}
}

@inproceedings{rao2021knowgnn,
  title={Know-GNN: Explainable Knowledge-Guided Graph Neural Network for Fraud Detection},
  author={Rao, Yizhuo and others},
  booktitle={ICONIP 2021},
  year={2021}
}

@inproceedings{zhou2024knowgraph,
  title={KnowGraph: Knowledge-Enabled Anomaly Detection via Logical Reasoning on Graph Data},
  author={Zhou, Andy and others},
  booktitle={Proceedings of ACM CCS},
  year={2024}
}

@inproceedings{visbeek2023dsc,
  title={Explainable Fraud Detection with Deep Symbolic Classification},
  author={Visbeek, Samantha and others},
  booktitle={XAI-FIN Workshop},
  year={2023}
}

@inproceedings{li2024sefraud,
  title={SEFraud: Graph-based Self-Explainable Fraud Detection via Interpretative Mask Learning},
  author={Li, Kaidi and others},
  booktitle={KDD},
  year={2024}
}

@inproceedings{pan2025huge,
  title={HUGE: Heterophily-Guided Unsupervised Graph Fraud Detection},
  author={Pan, Junjun and others},
  booktitle={AAAI},
  year={2025}
}

@inproceedings{park2022das,
  title={Domain Adaptation for Customs Fraud Detection via Prototype Sharing},
  author={Park, Sungwon and others},
  booktitle={AAAI},
  year={2022}
}

@article{knox2024drugbank,
  author  = {Knox, Craig and Wilson, Mike and Klinger, Christen M. and et al.},
  title   = {DrugBank 6.0: the DrugBank Knowledgebase for 2024},
  journal = {Nucleic Acids Research},
  year    = {2024},
  volume  = {52},
  number  = {D1},
  pages   = {D1265–D1275},
  doi     = {10.1093/nar/gkad976},
}

@article{cheng2019network,
  author  = {Cheng, Feixiong and Kov{\'a}cs, Ist{\'a}nv{\'a}n A. and Barab{\'a}si, Albert-L{\'a}szl{\'o}},
  title   = {Network-based prediction of drug combinations},
  journal = {Nature Communications},
  volume  = {10},
  number  = {1197},
  year    = {2019},
  doi     = {10.1038/s41467-019-09186-x},
  url     = {https://doi.org/10.1038/s41467-019-09186-x},
}

@inproceedings{bauder2016outlier,
  author    = {R. A. Bauder and T. M. Khoshgoftaar},
  title     = {A Probabilistic Programming Approach for Outlier Detection in Healthcare Claims},
  booktitle = {Proceedings of the IEEE 15th International Conference on Machine Learning and Applications (ICMLA)},
  year      = {2016},
  pages     = {347--354},
  publisher = {IEEE}
}

@article{herland2018bigdata,
  author = {Herland, Matthew and Khoshgoftaar, Taghi M. and Bauder, Russell A.},
  title = {Big Data Fraud Detection Using Multiple Medicare Data Sources},
  journal = {Journal of Big Data},
  volume = {5},
  pages = {29--35},
  year = {2018}
}

@inproceedings{sadiq2017primm,
  author = {Sadiq, S. and Tao, Y. and Yan, Y. and Shyu, M.-L.},
  title = {Mining Anomalies in Medicare Big Data Using Patient Rule Induction Method},
  booktitle = {Proceedings of the IEEE 3rd International Conference on Multimedia Big Data (BigMM)},
  year = {2017},
  pages = {185--192}
}

@article{johnson2019medicare,
  author = {Johnson, J. M. and Khoshgoftaar, T. M.},
  title = {Medicare Fraud Detection Using Neural Networks},
  journal = {Journal of Big Data},
  volume = {6},
  pages = {63--69},
  year = {2019}
}

@article{pang2021deep,
  author    = {Guansong Pang and Chunhua Shen and Longbing Cao and Anton van den Hengel},
  title     = {Deep Learning for Anomaly Detection: A Review},
  journal   = {ACM Computing Surveys},
  volume    = {54},
  number    = {2},
  pages     = {1--38},
  year      = {2021},
  publisher = {ACM}
}

@article{zafari2019topic,
  title={Topic modelling for medical prescription fraud and abuse detection},
  author={Zafari, Babak and Ekin, Tahir},
  journal={Journal of the Royal Statistical Society Series C: Applied Statistics},
  volume={68},
  number={3},
  pages={751--769},
  year={2019},
  publisher={Oxford University Press}
}

@inproceedings{zhao2024weakly,
  title={Weakly supervised anomaly detection via knowledge-data alignment},
  author={Zhao, Haihong and Zi, Chenyi and Liu, Yang and Zhang, Chen and Zhou, Yan and Li, Jia},
  booktitle={Proceedings of the ACM Web Conference 2024},
  pages={4083--4094},
  year={2024}
}

@article{suesserman2023procedure,
  title={Procedure code overutilization detection from healthcare claims using unsupervised deep learning methods},
  author={Suesserman, Michael and Gorny, Samantha and Lasaga, Daniel and Helms, John and Olson, Dan and Bowen, Edward and Bhattacharya, Sanmitra},
  journal={BMC Medical Informatics and Decision Making},
  volume={23},
  number={1},
  pages={196},
  year={2023},
  publisher={Springer}
}

@misc{fbi_healthcare_fraud,
  title={Health Care Fraud},
  author={{Federal Bureau of Investigation}},
  howpublished={\url{https://www.fbi.gov/investigate/white-collar-crime/health-care-fraud}},
  note={Accessed: 2025-05-20}
}

@article{shrank2019waste,
  title={Waste in the US Health Care System: Estimated Costs and Potential for Savings},
  author={Shrank, William H and Rogstad, Todd L and Parekh, Natasha},
  journal={JAMA},
  volume={322},
  number={15},
  pages={1501--1509},
  year={2019},
  publisher={American Medical Association},
  doi={10.1001/jama.2019.13978}
}

@misc{cmsPartD2025,
  title = {Medicare Part D Prescribers by Provider and Drug},
  author = {{Centers for Medicare \& Medicaid Services}},
  year = {2025},
  howpublished = {\url{https://data.cms.gov/provider-summary-by-type-of-service/medicare-part-d-prescribers/medicare-part-d-prescribers-by-provider-and-drug}},
  note = {Accessed: 2025-05-22}
}

@misc{leie2025,
  title = {List of Excluded Individuals/Entities (LEIE)},
  author = {{U.S. Department of Health and Human Services, Office of Inspector General}},
  year = {2025},
  howpublished = {\url{https://oig.hhs.gov/exclusions/exclusions_list.asp}},
  note = {Accessed: 2025-05-22}
}

@article{guy2017vitalsigns,
  title={Vital signs: Changes in opioid prescribing in the United States, 2006–2015},
  author={Guy, Gery P and Zhang, Kun and Bohm, Michele K and Losby, Jan and Lewis, Bryan and Young, Randall and Murphy, Laura B and Dowell, Deborah},
  journal={MMWR. Morbidity and Mortality Weekly Report},
  volume={66},
  number={26},
  pages={697},
  year={2017},
  publisher={Centers for Disease Control and Prevention}
}

@article{kennedy2025unsupervised,
  title={Unsupervised Feature Selection and Class Labeling for Credit Card Fraud},
  author={Kennedy, Robert K. L. and Nkole, Raymond S. and Mgutshini, Lucky N.},
  journal={Journal of Big Data},
  volume={12},
  number={1},
  pages={75},
  year={2025},
  publisher={Springer},
  doi={10.1186/s40537-025-01154-1},
  url={https://journalofbigdata.springeropen.com/articles/10.1186/s40537-025-01154-1}
}

@article{knight2008sinkhorn,
  title={The Sinkhorn--Knopp algorithm: convergence and applications},
  author={Knight, Philip A},
  journal={SIAM Journal on Matrix Analysis and Applications},
  volume={30},
  number={1},
  pages={261--275},
  year={2008},
  publisher={SIAM}
}

@article{sun2020medical,
  title={Medical Knowledge Graph to Enhance Fraud, Waste, and Abuse Detection on Claim Data: Model Development and Performance Evaluation},
  author={Sun, Haixia and Xiao, Jin and Zhu, Wei and He, Yilong and Zhang, Sheng and Xu, Xiaowei and Hou, Li and Li, Jiao and Ni, Yuan and Xie, Guotong},
  journal={JMIR Medical Informatics},
  volume={8},
  number={7},
  pages={e17653},
  year={2020},
  publisher={JMIR Publications},
  doi={10.2196/17653}
}

@article{zare2016improving,
  title={Improving Fraud and Abuse Detection in General Physician Claims: A Data Mining Study},
  author={Zare, Hamidreza and Ghasemi, Reza and Ghazanfari, Mehdi and Roshani, Fatemeh and Roshani, Mehdi},
  journal={JMIR Medical Informatics},
  volume={4},
  number={1},
  pages={e2},
  year={2016},
  publisher={JMIR Publications},
  doi={10.2196/medinform.4732}
}

@article{darwiche2002knowledge,
  title={A knowledge compilation map},
  author={Darwiche, Adnan and Marquis, Pierre},
  journal={Journal of Artificial Intelligence Research},
  volume={17},
  pages={229--264},
  year={2002}
}

@article{kipf2016semi,
  title={Semi-supervised classification with graph convolutional networks},
  author={Kipf, Thomas N and Welling, Max},
  journal={arXiv preprint arXiv:1609.02907},
  year={2016}
}

@article{Keiser2009,
  author  = {Michael J. Keiser and Vincent Setola and John J. Irwin and Christian Laggner and Atheir I. Abbas and Sandra J. Hufeisen and Niels H. Jensen and Michael B. Kuijer and Roberto C. Matos and Thuy B. Tran and Ryan Whaley and Richard A. Glennon and J{\'e}r{\^o}me Hert and Kelan L. H. Thomas and Douglas D. Edwards and Brian K. Shoichet and Bryan L. Roth},
  title   = {Predicting new molecular targets for known drugs},
  journal = {Nature},
  year    = {2009},
  volume  = {462},
  number  = {7270},
  pages   = {175--181},
  doi     = {10.1038/nature08506},
  pmid    = {19881490},
  pmcid   = {PMC2784146},
  url     = {https://pmc.ncbi.nlm.nih.gov/articles/PMC2784146/}
}

@article{Wang2013,
  author  = {Yong{-}Cui Wang and Shi{-}Long Chen and Nai{-}Yang Deng and Yong Wang},
  title   = {Network predicting drug's anatomical therapeutic chemical code},
  journal = {Bioinformatics},
  year    = {2013},
  volume  = {29},
  number  = {10},
  pages   = {1317--1324},
  doi     = {10.1093/bioinformatics/btt158},
  pmid    = {23564845},
  url     = {https://academic.oup.com/bioinformatics/article/29/10/1317/260431}
}

@article{Campillos2008,
  author  = {Monica Campillos and Michael Kuhn and Anne{-}Claude Gavin and Lars Juhl Jensen and Peer Bork},
  title   = {Drug target identification using side-effect similarity},
  journal = {Science},
  year    = {2008},
  volume  = {321},
  number  = {5886},
  pages   = {263--266},
  doi     = {10.1126/science.1158140},
  pmid    = {18621671},
  url     = {https://pubmed.ncbi.nlm.nih.gov/18621671/}
}

@techreport{OSC2022BrandDrugs,
  author       = {{Office of the New York State Comptroller}},
  title        = {Medicaid Program: Improper Payments for Brand Name Drugs},
  institution  = {Division of State Government Accountability, Office of the New York State Comptroller},
  number       = {Report 2020-S-62},
  address      = {Albany, NY},
  year         = {2022},
  month        = dec,
  url          = {https://www.osc.ny.gov/files/state-agencies/audits/pdf/sga-2023-20s62.pdf},
  urldate      = {2025-08-26}
}

@article{Dowell2023FCA,
  author       = {Dowell, Michael A.},
  title        = {Ruling Increases Pharmacy False Claims Act Risks},
  journal      = {U.S. Pharmacist},
  year         = {2023},
  volume       = {48},
  number       = {9},
  pages        = {7--12},
  month        = sep,
  url          = {https://www.uspharmacist.com/article/ruling-increases-pharmacy-false-claims-act-risks},
  urldate      = {2025-08-26}
}

@techreport{CMS2017Opioids,
  author       = {Centers for Medicare \& Medicaid Services and Healthcare Fraud Prevention Partnership and NORC at the University of Chicago},
  title        = {Healthcare Payer Strategies to Reduce the Harms of Opioids: White Paper},
  institution  = {U.S. Department of Health \& Human Services},
  year         = {2017},
  url          = {https://www.cms.gov/files/document/download-reducing-harms-opioids-white-paper.pdf}
}
\end{document}